\documentclass[letterpaper]{article} 
\usepackage{aaai23}  
\usepackage{times}  
\usepackage{helvet}  
\usepackage{courier}  
\usepackage[hyphens]{url}  
\usepackage{graphicx} 
\usepackage{natbib}  
\usepackage{caption} 
\usepackage{algorithm}
\usepackage{algorithmic}
\usepackage{color}
\usepackage{multirow}
\usepackage{diagbox}
\usepackage{comment}
\usepackage{bm}
\usepackage{amsfonts, amsmath, amsthm}

\usepackage{afterpage}
\usepackage{placeins}

\usepackage{flafter}
\usepackage{float}
\usepackage{capt-of}
\usepackage{subcaption}

\newtheorem{lemma}{Lemma}

\usepackage{newfloat}
\usepackage{listings}
\usepackage{algorithm}

\DeclareCaptionStyle{ruled}{labelfont=normalfont,labelsep=colon,strut=off} 
\floatstyle{ruled}
\newfloat{listing}{tb}{lst}{}
\floatname{listing}{Listing}
\title{RedCore: Relative Advantage Aware Cross-modal Representation Learning \\ for Missing Modalities with Imbalanced Missing Rates}
\author{
    Jun Sun\textsuperscript{\rm 1}, Xinxin Zhang\textsuperscript{\rm 2}, Shoukang Han\textsuperscript{\rm 1}, Yu-Ping Ruan\textsuperscript{\rm 1}, Taihao Li\textsuperscript{\rm 1}
}
\affiliations{
    \textsuperscript{\rm 1}Institute of Artificial Intelligence, Zhejiang Lab, Hangzhou, China\\
    \textsuperscript{\rm 2}Hangzhou Institute for Advanced Study, University of Chinese Academy of Sciences, Hangzhou, China \\
    sunjun16sj@gmail.com, zhangxinxin221@mails.ucas.ac.cn, \{hanshoukang, ypruan, lith\}@zhejianglab.com

}

\usepackage{bibentry}

\begin{document}

\maketitle

\begin{abstract}
Multimodal learning is susceptible to modality missing, which poses a major obstacle for its practical applications and, thus, invigorates increasing research interest. In this paper, we investigate two challenging problems: 1) when modality missing exists in the training data, how to exploit the incomplete samples while guaranteeing that they are properly supervised? 2) when the missing rates of different modalities vary, causing or exacerbating the imbalance among modalities, how to address the imbalance and ensure all modalities are well-trained? 
 To tackle these two challenges, we first introduce the variational information bottleneck (VIB) method for the cross-modal representation learning of missing modalities, which capitalizes on the available modalities and the labels as supervision. Then, accounting for the imbalanced missing rates, we define relative advantage to quantify the advantage of each modality over others. Accordingly, a bi-level optimization problem is formulated to adaptively regulate the supervision of all modalities during training. 
As a whole, the proposed approach features \textbf{Re}lative a\textbf{d}vantage aware \textbf{C}ross-m\textbf{o}dal \textbf{r}epresentation l\textbf{e}arning (abbreviated as \textbf{RedCore}) for missing modalities with imbalanced missing rates. Extensive empirical results demonstrate that RedCore outperforms competing models in that it exhibits superior robustness against either large or imbalanced missing rates.

\end{abstract}

\section{Introduction}
Multimodal learning leverages  heterogeneous and comprehensive signals, such as acoustic, visual, lexical information to perform typical machine learning tasks, for instance, clustering, regression, classification, and retrieval \cite{9113457, 10004588, han2022trusted}. Relative to its unimodal counterpart, multimodal learning has demonstrated great success in numerous applications, including but not limited to medical analysis \cite{liu2023m3ae}, action recognition \cite{woo2023towards}, affective computing\cite{sun2023layer}. Nevertheless, multimodal learning is inevitably faced with the modality missing issue due to malfunctioning sensors, high data acquisition costs, privacy concerns, etc. In this circumstance, a model trained with full modalities can be highly sensitive to modality missing in inference phase, and thus suffers from considerable performance deterioration in practical scenarios where some modalities are unavailable. In order to guarantee the reliability and safety in practice, to develop multimodal models robust against modality missing has raised widespread interest from industry and research communities \cite{tang2022deep}. 

In the presence of missing modalities, popular studies usually first conduct missing modality imputation before further computations. Naive zero or mean value \cite{wen2021unified, hu2018doubly} padding is of minimal exertion, yet can be quite inaccurate. 
More advanced statistical techniques primarily rely on the nonparametric Bayesian approach \cite{manrique2017bayesian}, k-nearest neighbors \cite{jadhav2019comparison}, Gaussian processes \cite{jafrasteh2023gaussian}. 
Current prevalent models building upon deep learning are more capable of exploring the correlation between modalities to impute the missing ones. 
It is straightforward that generative methods like generative adversarial network \cite{yoon2020gamin} and variational autoencoder \cite{rezaei2021learning}, can be applied to the reconstruction of missing modalities.
Besides, knowledge distillation can also be employed to predict the information of missing modalities from a large model pre-trained with large corpora of modality-complete data \cite{wang2020multimodal, wei2023mmanet}.

\begin{figure}[t]
	\centering
	\includegraphics[width=1\linewidth]{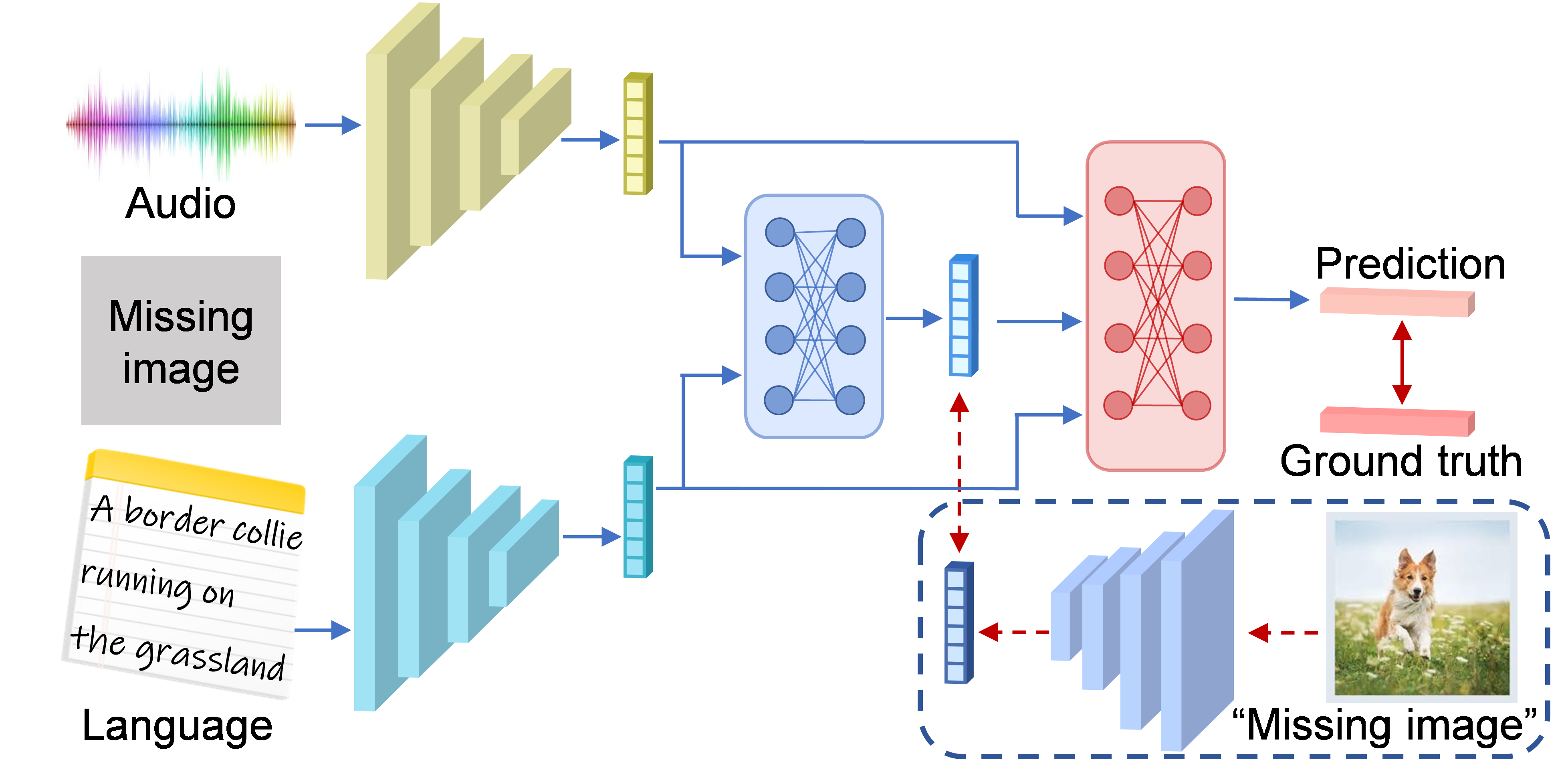}   
	\vspace{-0.6cm}
		\caption{The typical architecture of PDT. With the image module in the dashed blue box removed, it corresponds to IDT. The bi-directional arrows represent supervision.}
	\label{fig:PDTvsIDT}
	\vspace{-0.1cm}
\end{figure}

According to whether all collected data is utilized for training, current schemes can be categorized into two regimes: 1) perfect data training (PDT) \cite{zhao2021missing, woo2023towards}, where samples with missing modalities are discarded, and only the complete samples are used for model training; and 2) imperfect data training (IDT)\cite{ma2021smil}, where samples with missing modalities are retained for training together with complete samples. In PDT, some modalities of the training samples are masked to simulate the absence of modalities, which enforces the model to be aware of the situation of modality missing; although the masked modalities do not contribute directly to the forward feature extraction, they play a role in supervising the learning procedure, as illustrated in Figure~\ref{fig:PDTvsIDT}. In IDT, albeit the missing modalities of imperfect samples are absolutely inaccessible, their available modalities carrying useful information can facilitate the model training. In comparison, the obvious merit of IDT lies in that it is more data-efficient than PDT, as all collected data is fully exploited. However, the downside is that the model design and training of IDT is more challenging, since the lack of supervision for the imputation can negatively impact the model. 
To sum up, the dilemma of prior works is that they have to compromised between data efficiency and supervision sufficiency, which is the first motivation of the present work.

Furthermore, in practical scenarios, the missing rates of different modalities can vary significantly. For instance, it is likely that 80\% of all training samples might miss one modality, while only 20\% might miss another modality. Consequently , this can lead to modality imbalance, where weak modalities (those with high missing rates in our case) might be dominated by strong ones, suffering from being under-trained. Modality imbalance problem is inherent for multimodal learning, since some modalities can naturally perform the task better and learn faster than others. Despite the efforts of previous research to address the modality imbalance issue \cite{sun2021learning, peng2022balanced, wang2020makes, wu2022characterizing}, it requires special attention in the context of modality missing. There are two reasons: firstly, modality imbalance can be aggravated by imbalanced missing rates;
secondly, apart from the final task, imputation is also influenced by imbalance missing rates.
To the best of our knowledge, no existing relevant study has taken the imbalanced missing rates across modalities into account, which constitutes our second motivation.

With the above analysis, our work aims to achieve two primary goals: 1) to develop a model that enjoys the data efficiency and supervision sufficiency simultaneously for multimodal learning with modality missing; and 2) to propose a training scheme to solve the problem of imbalance missing rates. For the model design and training, we adopt the IDT regime, which, as mentioned above, is data efficient. Taking advantage of the variational information bottleneck (VIB) method \cite{alemi2016deep, bang2021explaining}, we devise a model with cross-modal representation learning for missing modality imputation. With VIB, each modality learns an optimal representation that retains minimal information from the original feature while capturing the maximal information relevant to the task. The novelty of our method is that for each modality, regardless of being missing or available, a cross-modal representation generated from the optimal representation of all other available modalities is learned. The supervision of the cross-modal representation learning for available modalities is from both the features and the labels, and for missing modalities it comes from only the labels. 

In order to balance the modalities, we first define the relative advantage of each individual modality according to the imputation loss during training. Subsequently, with relative advantage, a bi-level optimization problem \cite{ji2021bilevel, chen2023gradient} is formulated to regulate the supervisions of different modalities. Based on the analysis of the bi-level optimization problem we propose a two-double algorithm, corresponding to the joint training and adaptive supervision regulation approach.

In summary, we propose a novel approach --- \textbf{Re}lative a\textbf{d}vantage aware \textbf{C}ross-m\textbf{o}dal \textbf{r}epresentation l\textbf{e}arning, referred to as \textbf{RedCore}, which addresses the challenge of missing modalities with imbalanced missing rates. The contributions are primarily threefold:

\begin{itemize}
	\item Building upon the VIB method, we leverage the available modalities and the labels as supervision for cross-modal representation learning, guaranteeing that the imputation modules are sufficiently supervised in the data-efficient IDT training framework.
	\item We propose the concept of relative advantage for modalities, according to which a bi-level optimization problem is formulated to adaptively balance their training. 
	\item Numerical results validate the effectiveness of RedCore and demonstrate its superior performance compared with state-of-the-art models.
\end{itemize}

\section{Related Works}
\subsection{Missing Modality Imputation}
Imputation is of central importance in the literature addressing modality missing. The vast majority of existing works fall in the category of implicit imputation (discriminative approach) and explicit imputation (generative approach) \cite{morales2022simultaneous, miyaguchi2022variational}. 
The former carries out imputation implicitly or simultaneously with the prediction \cite{  josse2020consistency,jeong2022fairness}. 
The latter typically maintains an explicit generative module for missing value completion, which spans a wide spectrum of schemes, including kernel learning, graph learning, matrix factorization, and subspace learning. 
Kernel learning is usually adopted in clustering task with incomplete modalities, where the clustering and kernel feature learning can be performed alternatively or simultaneously \cite{liu2018late, liu2020efficient}. 
For graph based approaches, subgraph of each accessible modality is usually pre-constructed, and cross-modal graph learning is proposed to capture the missing local graph structure and complete the global graph \cite{li2022refining}. 
Matrix factorization can be utilized to obtain the latent factors of missing modalities by leveraging complementary multimodal information and underlying cross-modal relation \cite{liu2021novel}.
Subspace learning maps the multimodal features into low-dimensional latent representations, and exploit the correlation between modalities to impute the missing latent representation. A unified subspace embedding alignment framework is proposed in \cite{liu2020efficient}, where the missing modalities are reconstructed with Laplacian regularized error matrices compensation. Besides, autoencoder based methods are representative subspace learning approaches for missing modality generation \cite{du2018semi}.
Via stacking a series of residual autoencoders, cascaded residual autoencoder (CRA) is designed to promote the missing modality modeling ability of autoencoders \cite{tran2017missing}. Additionally, incorporating cycle consistency loss to CRA can further improve the performance of cross-modal imputation \cite{zhao2021missing}.

\subsection{Fairness and Imbalance in Multimodal Learning}
Because of the modality discrepancy, multimodal learning naturally encounters the concern of fairness and imbalance \cite{zhang2021assessing}, which is highlighted in context of modality incompleteness.
The fairness of machine learning methods with missing attributes are comprehensively investigated in \cite{fernando2021missing}. A reweighting scheme that assigns lower weight to data sample with missing values to promote fairness is proposed in \cite{wang2021analyzing}. A decision tree without explicit imputation is trained by optimizing a fairness-regularized objective function in \cite{jeong2022fairness}. Another line of works that inspire our proposed approach to addressing the modality imbalance are those coping with the imbalanced convergence rates of different modalities in multimodal model training. The learning rates of different modalities are adaptively tuned during training according to the loss, the overfitting estimation, and the parameters \& their gradients of each modality in works \cite{sun2021learning, peng2022balanced}, \cite{wang2020makes} and \cite{wu2022characterizing}, respectively.
Our work distinguishes from these prior studies in two aspects: 1) we particularly focus on balancing the training of representation imputation; and 2) in stead of balancing the modalities heuristically, we formulate an optimization problem for supervision regulation.

\section{Method: RedCore}
The proposed approach, RedCore, is composed of two components --- Core and Red, as elaborated in the following subsections. Prior to that, we give some notation definitions.

\subsubsection{Notations:} Suppose a multimodal training dataset contains $N$ samples, each with $M$ modalities. For ease of expression, let us define an auxiliary modality as a union of all modalities, and thus the total number of modalities is $M+1$. Let $[P]$ for any positive integer $P$ denote the set $\{1, 2, \cdots, P\}$. The training samples are denoted by $(\{\bm{x}^m_n\}_{m\in[M]}, \{y_n^m\}_{m \in[M+1]})$, where $n\in[N]$ indexes the samples, $\bm{x}^m_n\in\mathbb{R}^{d_{m}}$ represents the $d_m$-dimensional feature vector (the feature can also be vector sequence) of modality $m, \forall m\in [M]$, and $y_n^m$ represents the label corresponding to modality $m, \forall m \in[M+1]$ (for datasets where all modalities share a common label, $y_n^1 = y_n^2 = \cdots = y_n^{M+1}$ holds). For the consistency of expression, we use $\bm{x}_n^{M+1}:=[\bm{x}_n^1; \bm{x}_n^2;\cdots;\bm{x}^M_n]$ to collect all features of sample $n$. Let matrix $\bm{C}\in\mathbb{R}^{N\times M}$ be the modality presence indication matrix, and let $C_{nm}$ denote the $n,m$-th entry of $\bm{C}$. Then, $C_{nm}=1$, if modality $m$ of sample $n$ is available; otherwise, $C_{nm}=0$. For missing modalities, their labels are endowed with the joint label; namely, for any sample $n, n\in[N]$ with modality missing, $y_n^m = y_n^{M+1}$ holds for all $m\in\{m\mid C_{nm}=0\}$.
The missing rate of modality $m, \forall m\in [M]$, $MR^m$, is calculated as $MR^m =(N- \sum_{n\in[N]} C_{nm})/N$. We assume that $MR^m\in[0,1)$ to ensure that all modalities exist in the training dataset. 

Let $\bm{x}^m$ and $y^m, \forall m \in[M+1]$ respectively denote the general random variables of the feature and label, and $\bm{x}^m_n$ and $y_n^m$ are realizations of them, respectively. Let $\bm{z}^m\in \mathbb{R}^d$, a map of $\bm{x}^m$, denote the representation of modality $m$. Similarly, $\bm{z}_n^m$ is a realization of $\bm{z}^m$ (for brevity, we assume the representations of all modalities are $d$-dimensional vectors). 

\begin{figure}[t]
	\centering

		\centering
		\includegraphics[width=2.2in]{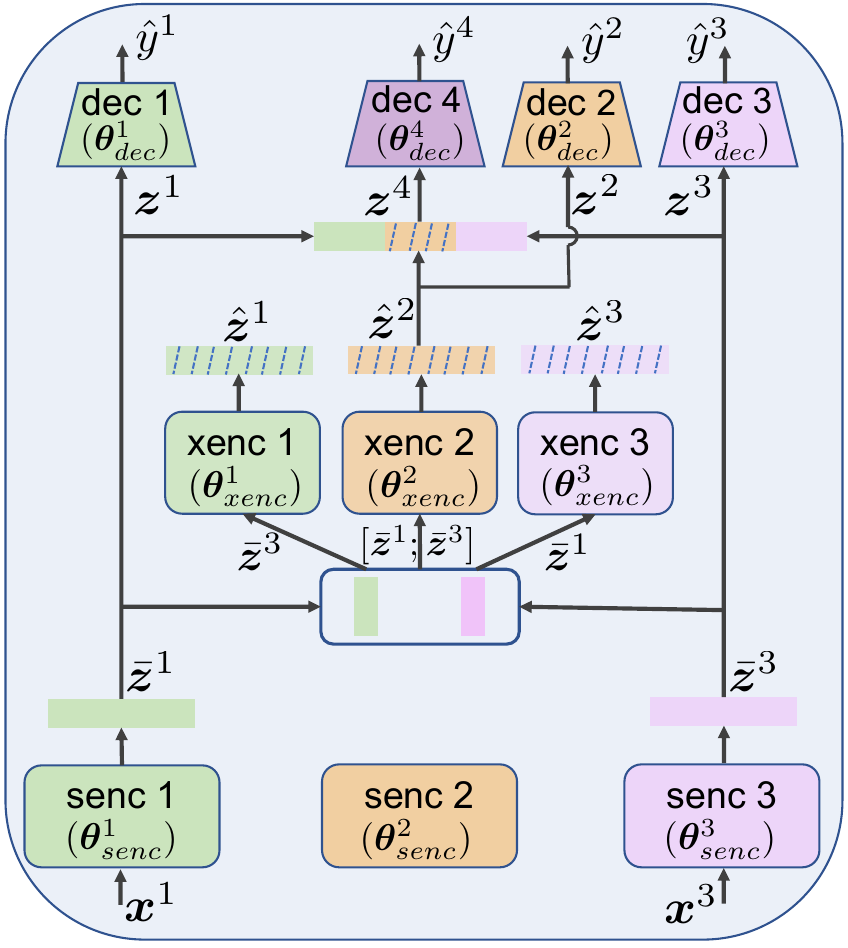}

	\vspace{-0.2cm}
	\caption{Model architecture (with 3 modalities as an example). Training on a sample with modality 2 missing. }
   \label{fig:traintest}
	\vspace{-0.1cm}
\end{figure}

\subsection{Core --- Cross Modal Representation Learning}

In this subsection, we will introduce the framework of our model and VIB based representation learning.

\vspace{0.15cm}
\noindent{\textbf{A. Model architecture}}
\vspace{0.1cm}

\noindent The proposed model architecture (with 3 modalities as an example) is visualized in Figure~\ref{fig:traintest},  depicting the training computation flow on a sample with the second modality missing. In specific, the feature of each available modality $\bm{x}^m$ is mapped to its representation $\bar{\bm{z}}^m$ via a variational encoder with parameter $\bm{\theta}_{senc}^m$ (senc: self-modal encoder). Subsequently, through cross-modal encoder with parameter $\bm{\theta}_{xenc}^m$ (xenc: cross-modal encoder), each modality obtains a representation $\hat{\bm{z}}^m$  from  the representation $\bar{\bm{z}}^m$ of all other available modalities. The final representation $\bm{z}^m$ is attained as $\bar{\bm{z}}^m$ (resp., $\hat{\bm{z}}^m$) if modality $m$ is present (resp., missing). Mathematically, for feature $\bm{x}_n^m$, its representation $\bm{z}_n^m$ is derived as following. Firstly, the representations of available modalities are sampled from distribution $p(\bar{\bm{z}}^m|\bm{x}^m; \bm{\theta}^m_{senc})$, i.e., $\bar{\bm{z}}^m_n \sim p(\bar{\bm{z}}^m|\bm{x}^m; \bm{\theta}^m_{senc}), \forall m\in\{m\mid C_{nm} = 1\}$. Then, the cross-modal representations are calculated for all modalities as $\hat{\bm{z}}_n^m = g(\{\bar{\bm{z}}_n^{m'}\}_{m'\in\mathcal{A}_n^m}; \bm{\theta}_{xenc}^m)$, where $\mathcal{A}_n^m$ is defined as the set of all available modalities other than modality $m$ of sample $n$, namely, $\mathcal{A}_n^m := \{m'\in[M]\mid C_{nm'} = 1, m'\neq m\}$. Finally, with the above notations, the ultimate representation of feature $\bm{x}_n^m, \forall n\in[N], \forall m\in[M]$ boils down to: 
\begin{equation}\label{eq:z}
	    \bm{z}_n^m= 
	\begin{cases}
		\bar{\bm{z}}^m_n, & \text{if} ~~ C_{nm} = 1;\\
		\hat{\bm{z}}^m_n, & \text{otherwise.}
	\end{cases}
\end{equation}
The representation of modality $M+1$ is the concatenation of $z_n^m$ for all $m \in [M]$, i.e., $\bm{z}^{M+1}_n := [\bm{z}_n^1; \bm{z}_n^2; \cdots; \bm{z}_n^M]$. When we use $\bm{\theta}_{enc}^m :=\{\bm{\theta}_{senc}^m, \bm{\theta}_{xenc}^m\}$ to collect encoder parameters for modality $m$, and $\bm{\theta}_{enc}^{M+1}:= \{\bm{\theta}_{enc}^m\}_{m \in[M]}$ to gather all encoder parameters, the above representation can be summarized as $\bm{z}_n^m\sim p(\bm{z}^m|\bm{x}^m; \bm{\theta}_{enc}^m), \forall m \in [M+1]$.

For a typical multimodal learning setting, the joint representation $\bm{z}^{M+1}_n$ is used for the final task, which is tantamount to passing the representation to a decoder with parameter $\bm{\theta}^{M+1}_{dec}$ and yielding the label prediction: $\hat{y}_n^{M+1}\sim p(y^{M+1}|\bm{z}^{M+1};\bm{\theta}_{dec}^{M+1})$. In addition, in our framework, each individual modality also performs label prediction, even if all modalities share the same label ($y_n^1 \!=\! y_n^2 \!\!=\!\! \cdots \! = \!\! y_n^{M+1}$). This relies on the decoder $\bm{\theta}_{dec}^m$: $y^m_n\sim p(y^{m}|\bm{z}^m;\bm{\theta}_{dec}^{m}), m\in[M]$. 
The reason why these additional predictions are conducted will be given in the next subsection. 

\vspace{0.15cm}
\noindent{\textbf{B. VIB based representation learning}}
\vspace{0.1cm}

\noindent The goal of variational information bottleneck based learning is to obtain a representation that is maximally informative about the target and maximally compressive about the raw feature. In our scenario, this can be cast as the following optimization problem which minimizes the information bottleneck (IB) loss.
\begin{equation*}
\min\limits_{\bm{\theta}} ~ \mathcal{L}_{I\!B}(\bm{\theta}) :=\!\! \sum\limits_{m\in[M+1]} \!\!\! I(\bm{x}^m, \bm{z}^m;\bm{\theta}) - \gamma I(\bm{z}^m, y^m; \bm{\theta}),
\end{equation*}
where $\bm{\theta} := \{\bm{\theta}^m_{enc}, \bm{\theta}^m_{dec}\}_{m \in[M+1]}$ collects all model parameters, $I(\cdot, \cdot)$ represents the mutual information between two random variables, and the coefficient $\gamma>0$ is a predefined constant that balances the two terms in the loss function. 

Since it is intractable to compute the mutual information $I(\bm{x}^m, \bm{z}^m;\bm{\theta})$ and $I(\bm{z}^m, y^m;\bm{\theta})$, the variational upper bound of $\mathcal{L}_{I\!B}(\bm{\theta})$ is invoked \cite{alemi2016deep}:
\begin{equation} \label{eq:LVIB}
	\begin{aligned}
	\mathcal{L}_{I\!B}(\bm{\theta}) &\leq \mathcal{L}_{V\!I\!B}(\bm{\theta}) \\& := \sum\limits_{m=1}^{M+1} \mathbb{E}[  D_{K\!L}(p(\bm{z}^m|\bm{x}^m;\bm{\theta}^m_{enc})||q(\bm{z}^m)) \\
	& ~~~~~~~~~~~~~~~~~~~ - \gamma \log q(y^m|\bm{z}^m; \bm{\theta}^m_{dec})] ,
	\end{aligned}
\end{equation}
where the expectation $\mathbb{E}[\cdot]$ is taken with respect to random variables $\bm{x}^m, \bm{z}^m, y^m$; $D_{K\!L}(\cdot||\cdot)$ denotes the KL-divergence between two distributions; $q(\bm{z}^m)$ and $q(y^m|\bm{z}^m; \bm{\theta}^m_{dec})$ denote variational distributions approximating $p(\bm{z}^m|\bm{x}^m; \bm{\theta}^m_{enc})$ and $p(y^m|\bm{z}^m; \bm{\theta}^m_{dec})$, respectively.

When $p(\bm{z}^m|\bm{x}^m; \bm{\theta}^m_{enc})$ and $q(\bm{z}^m))$ take the Gaussian form, applying the reparameterization trick \cite{kingmaauto} to sample $\bm{z}^m$ results in the following approximation of $\mathcal{L}_{V\!I\!B}(\bm{\theta})$ \cite{alemi2016deep}. (The details are included in the supplementary materials for completeness.)
 \begin{equation}\label{eq:LVIBapp}
	\begin{aligned}
		\!\!\! \tilde{\mathcal{L}}_{V\!I\!B}(\bm{\theta}) =  & \frac{1}{N} \sum\limits_{n=1}^N\sum\limits_{m=1}^{M+1} D_{K\!L}(p(\bm{z}_n^m|\bm{x}_n^m;\bm{\theta}^m_{enc})||q(\bm{z}_n^m)) \\
		&  ~~~~~~~~~~~~~~~~~~~~ -\gamma \mathbb{E}_{\bm{\epsilon}}[\log q(y_n^m|\bm{z}_n^m; \bm{\theta}^m_{dec})],
	\end{aligned}
\end{equation}
where $\bm{\epsilon}\!\sim\! \mathcal{N}(\bm{0}_d, \bm{I}_d)$ is the random variable from the reparameterization trick. This formulation circumvents the expectation in eq.\eqref{eq:LVIB}, and thus allows us to compute unbiased stochastic gradient via backpropagating of a single sample.

Minimizing $\tilde{\mathcal{L}}_{V\!I\!B}(\bm{\theta})$ leads to an optimal representation for each modality in the sense of information bottleneck theory, meaning compressing the feature and meanwhile maximizing the prediction accuracy, which corresponds respectively to the first and second terms in eq.~\eqref{eq:LVIBapp}. This explains why we enforce each modality to predict its own label --- to attain its optimal task-relevant representation. Consequently, the cross-modal learning module is freed from dealing with task-irrelevant details, which thereby relieves the difficulty in learning parameters $\bm{\theta}_{xenc}^m, \forall m\in[M]$.

\vspace{0.15cm}
\noindent \textbf{C. Cross-modal representation learning} 
\vspace{0.1cm}

\noindent The imputation in our framework translates to the cross-modal representation, which is learned for all modalities regardless of whether they are present or absent for each sample. For both missing and available  modalities, the cross-modal encoder receives supervision from the label, which can be achieved via minimizing the VIB loss $\tilde{\mathcal{L}}_{V\!I\!B}(\bm{\theta})$. For available modalities, the cross-modal encoder receives additional supervision from the feature by minimizing the following mean squared error (MSE), i.e., the imputation loss:
\begin{equation*}
	\mathcal{L}_{M\!S\!E}^m(\bm{\theta}) := \frac{1}{N_m}\sum\limits_{n=1}^N||\bm{z}_n^m - \hat{\bm{z}}_n^m||_2^2,
\end{equation*}
where $N_m = \sum_{n\in[N]} C_{nm}$ is the number of samples that have modality $m$. From the definition of $\bm{z}_n^m$ in eq.~\eqref{eq:z}, it can be observed that the imputation loss $\mathcal{L}_{M\!S\!E}^m(\bm{\theta})$ accounts only for the representation recovery loss of available modalities  (because $\bm{z}_n^m=\hat{\bm{z}}_n^m$ holds for missing modalities).

The overall loss function for training is 
\begin{equation*}
	\mathcal{L}(\bm{\theta}): = \tilde{\mathcal{L}}_{V\!I\!B}(\bm{\theta}) + \sum\limits_{m\in[M]} \eta^m \mathcal{L}_{M\!S\!E}^m(\bm{\theta}),
\end{equation*}
where the coefficients $\eta^m, \forall m\in[M]$ weigh the representation reconstruction losses of different modalities.

To summarize, the supervisions of the encoders, including both the self-modal representation encoder ($\bm{\theta}^m_{senc}, m\in[M]$) and the cross-modal representation encoder ($\bm{\theta}^m_{xenc}, m\in[M]$), come from the label and the available modalities. As Figure~\ref{fig:traintest}(b) illustrates, when training  on a certain sample with missing modality $m$, the parameter $\bm{\theta}^m_{senc}$ and $\bm{\theta}^m_{xenc}$ do not receive supervision from the feature, yet the latter are supervised by the label. Therefore, in the circumstance of imbalanced missing rates, the parameters corresponding to modalities with larger missing rates receive supervision less frequently. This can cause some modalities to be overfitting while others underfitting, which is the issue to be addressed in the following subsection.

\subsection{Red --- Relative Advantage Aware Supervision Regulation}

In this part, we start with defining the relative advantage of each modality to quantify how strong (or weak) it is compared to other modalities. Then, according to the relative advantage, the strength of supervision, $\eta^m$, is adaptively regulated to balance the learning process across all modalities. 

\vspace{0.15cm}
\noindent{\textbf{A. Relative advantage}}
\vspace{0.1cm}

\noindent The strong modalities are usually better at learning or have smaller missing rates, which effects relatively smaller imputation losses during training. This inspires us to define the relative advantage according to the imputation loss. Concretely, for any $\bm{\theta}$, the average imputation loss over modalities is first calculated as: 
\begin{equation} \label{eq:aveloss}
	\bar{\mathcal{L}}_{M\!S\!E}(\bm{\theta}) = \sum\limits_{m\in[M]} \mathcal{L}_{M\!S\!E}^m(\bm{\theta})/M.
\end{equation}
With $\bar{\mathcal{L}}_{MSE}^m(\bm{\theta})$, relative advantage is formally defined as: 
\begin{equation} \label{eq:ra}
	RA^m(\bm{\theta}):= [\bar{\mathcal{L}}_{M\!S\!E}(\bm{\theta}) - \mathcal{L}_{M\!S\!E}^m(\bm{\theta})]/\bar{\mathcal{L}}_{M\!S\!E}(\bm{\theta}).
\end{equation}
When $RA^m(\bm{\theta})>0$ (resp., $<0$), modality $m$ is considered strong (resp., weak) relative to other modalities; $RA^m(\bm{\theta})=0$ indicates that modality $m$ is at the balancing point. 

\begin{algorithm}[t!]
	\caption{ Dynamic supervision regulation}\label{alg:red}
	\begin{algorithmic}[1]
		\STATE \textbf{Initialization:} randomly choose $\bm{\eta}_0$ and $\bm{\theta}_0$. \
		\FOR {$s=0$ to $S-1$ }
		\STATE calculate $\bm{r}(\bm{\theta}_{sK})$ according to eqs.~\eqref{eq:mal}, \eqref{eq:aveloss} and \eqref{eq:ra};
		\STATE update $\bm{\eta}$ according to eq.~\eqref{eq:superupdate} with  $\bm{r}(\bm{\theta}_{sK})$.
		\FOR{$k=0$ to $K-1$}
		\STATE update $\bm{\theta}$ using stochastic gradient descent (or other schemes, e.g., Adam):
		\STATE $\bm{\theta}_{sK+k+1} = \bm{\theta}_{sK+k} - \alpha_2 \tilde{\nabla}\mathcal{L}(\bm{\theta}_{sK+k})$, where $\tilde{\nabla}\mathcal{L}(\bm{\theta}_{sK+k})$ is the stochastic gradient of $\mathcal{L}(\bm{\theta}_{sK+k})$.
		\ENDFOR
		\ENDFOR 
		\STATE \textbf{Return:} Model parameter $\bm{\theta}_{S\!K}$.
	\end{algorithmic}
\end{algorithm}

\vspace{0.15cm}
\noindent\textbf{{B. Relative advantage aware supervision regulation}}
\vspace{0.1cm}

\noindent For brevity, let $\bm{\eta} = [\eta^1, \eta^2, \cdots, \eta^M]^T$ and $\bm{r}(\bm{\theta}) = [RA^1(\bm{\theta}), RA^2(\bm{\theta}), \cdots, RA^M(\bm{\theta})]^T$ collect all the supervisions and relative advantages, respectively.
In order to balance the modalities, the strength of the supervision $\bm{\eta}$ should be negatively correlated with the relative advantage $\bm{r}(\bm{\theta})$. Therefore, the objective of supervision regulation is to minimize the correlation between $\bm{\eta}$ and $\bm{r}(\bm{\theta})$. The joint training and supervision regulation problem can be formulated as the following bi-level optimization problem for the non-trivial case where $\bm{r}(\bm{\theta})\neq \bm{0}$ ($\bm{0}$ is an all-zero vector of proper size):
\begin{subequations}\label{eq:bilevel}
	\vspace{-0.2cm}
	\begin{align}
		\min\limits_{\bm{\eta},\bm{\theta}}& ~~~  \bm{r}(\bm{\theta})^T\bm{\eta} \\
		\text{s.t.} & ~~~ \bm{\eta} \ge \xi_1 \bm{1} \label{eq:positive} \\
		& ~~~ ||\bm{\eta}||_p = \xi_2 \label{eq:ub} \\ 
		& ~~~ \bm{\theta} = \arg \min\limits_{\bm{\theta}}  \mathcal{L}_{V\!I\!B}(\bm{\theta}) + \sum\limits_{m\in[M]} \eta^m \mathcal{L}_{M\!S\!E}^m(\bm{\theta}) \label{eq:training}
	\end{align}
\end{subequations}
where $\xi_1>0$ is a small constant to guarantee that the supervision is positive ($\bm{1}$ is an all-one vector); constraint~\eqref{eq:ub} constrains the $\ell_p$-norm ($p\ge 1$) of the supervision by a given constant $\xi_2$, and guarantees the stability of the training; constraint~\eqref{eq:training} which involves the model training given the supervision is the lower level problem embedded in the upper one, eq. \eqref{eq:bilevel}. The upper level problem, with optimization variable $\bm{\eta}$, corresponds to the supervision regulation --- adaptively tuning the supervision via minimizing the correlation between the supervision and the relative advantage. 

\begin{figure*}
\begin{minipage}{\textwidth}
\begin{subfigure}[b]{0.33\linewidth} 
	\centering
	\includegraphics[height=1.725in]{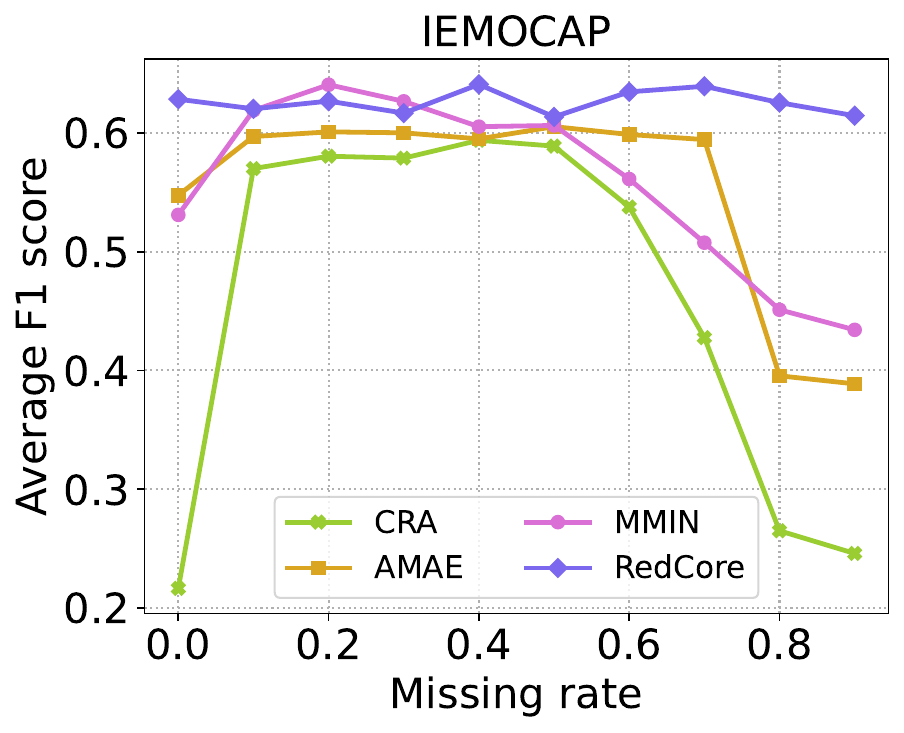}  
	\leftline{~~~(a) {\small F1} score vs missing rate on {\small IEMOCAP.}}\medskip
	\label{fig:side:a}
\end{subfigure}%
\begin{subfigure}[b]{0.34\linewidth} 
	\centering
	\includegraphics[height=1.725in]{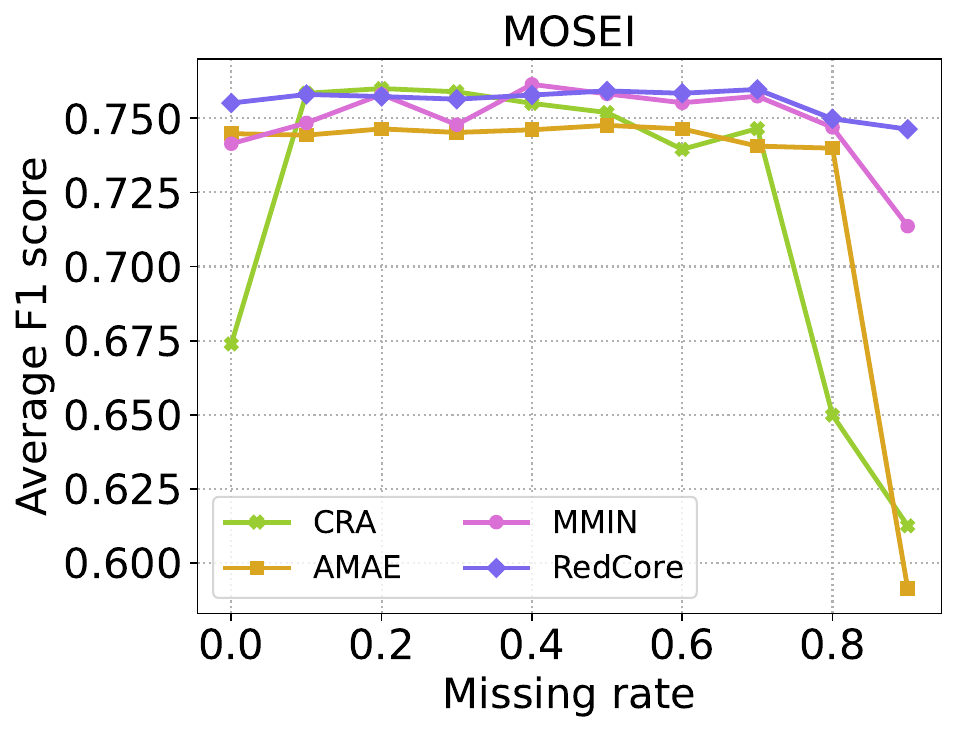}
	\centerline{~~~~~~~~(b) {\small F1} score vs missing rate on {\small MOSEI.}}\medskip
	\label{fig:side:b}
\end{subfigure}%
\begin{subfigure}[b]{0.33\linewidth} 
	\centering
	\includegraphics[height=1.725in]{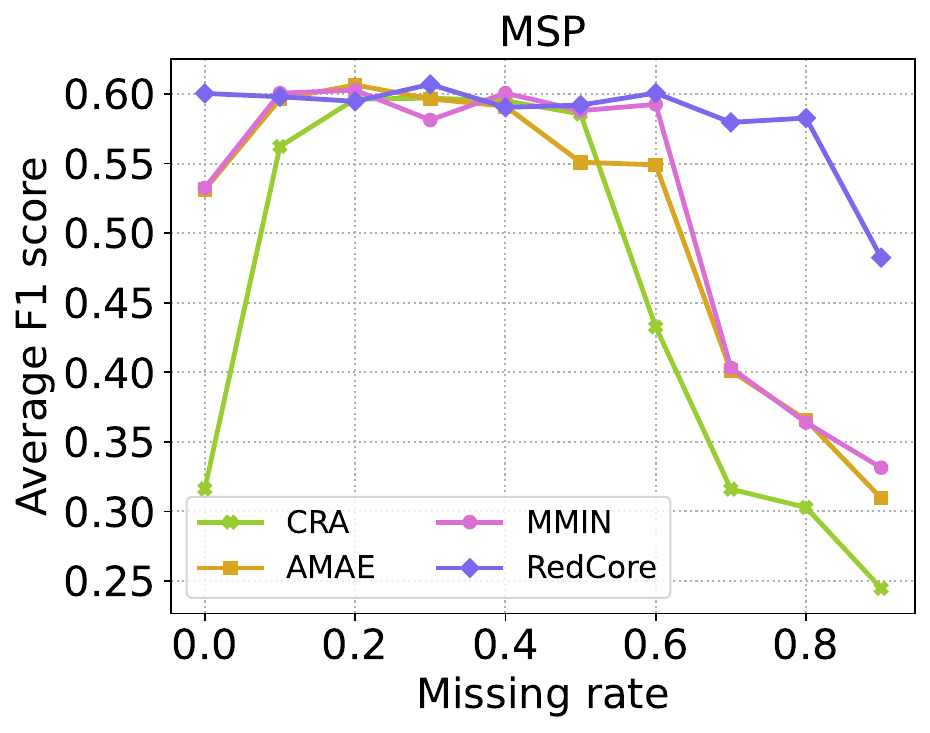}
	\centerline{~~~(c) {\small F1} score vs missing rate on {\small MSP.}}\medskip
	\label{fig:side:c}
\end{subfigure}%
\vspace{-0.4cm}
\caption{Average F1 score with varying missing rates for different datasets.}\label{fig:F1vsMR}
\vspace{-0.2cm}
\end{minipage}

\vspace{-0.1cm}
\end{figure*}

In the sequel, we will analyze the property of the above problem, and propose our algorithm accordingly.
A bi-level optimization problem is generally non-convex, which applies to problem~\eqref{eq:bilevel}. Nonetheless, an appealing characteristic of problem ~\eqref{eq:bilevel} holds, as is stated below. 

\begin{lemma} \label{lemma:convex}
	For any fixed parameter $\bm{\theta}$, problem~\eqref{eq:bilevel} is equivalent to the following convex problem. 
	\begin{subequations}\label{eq:bilevel2}
		\begin{align}
			\min\limits_{\bm{\eta}}& ~~~  \bm{r}(\bm{\theta})^T\bm{\eta} \\
			\text{s.t.} & ~~~ \bm{\eta} \ge \xi_1 \bm{1} \label{eq:positive2} \\
			& ~~~ ||\bm{\eta}||_p \le \xi_2 \label{eq:ub2}
		\end{align}
	\end{subequations}
\end{lemma}
Due to space limitations, the proof is delegated to the appendix in the supplementary materials. 
For a fixed parameter $\bm{\theta}$, it is obvious that constraint~\eqref{eq:training} can be directly eliminated. Then, the problem \eqref{eq:bilevel2} is an immediate result of relaxing the nonlinear equality constraint \eqref{eq:ub} to inequality \eqref{eq:ub2}.
	With Lemma~\ref{eq:bilevel2}, for a fixed parameter $\bm{\theta}$, we apply projected gradient descent to update $\bm{\eta}$ as the following eq.~\eqref{eq:superupdate}, for which the convergence is guaranteed \cite{vu2022asymptotic}. 
	\begin{equation} \label{eq:superupdate}
		\bm{\eta}_{s+1} = \textit{Pro}_{\Gamma_1}[\bm{\eta}_s - \alpha_1 \bm{r}(\bm{\theta}_s)],
	\end{equation}
	where $s$ indexes the updating step of $\bm{\eta}$, $\Gamma_1=\{\bm{\eta}\in\mathbb{R}^M\mid\bm{\eta} \!\ge \! \xi_1 \bm{1}$ , $||\bm{\eta}||_p = \xi_2\}$ is the feasible domain of $\bm{\eta}$, $\textit{Pro}_{\Gamma_1}[\cdot]$ represents projection to domain $\Gamma_1$, and $\bm{r}(\bm{\theta}_s)$ is the gradient of the objective function $\bm{r}(\bm{\theta}_s)^T\bm{\eta}$ w.r.t. $\bm{\eta}$. It is straightforward from eq.~\eqref{eq:superupdate} that the supervision regulation amounts to reducing the strength of its supervision if modality $m (m\in[M])$ has advantage over others, i.e., $RA^m(\bm{\theta})>0$; otherwise, the supervision strength should be increased.

\begin{table*}[t]
		\small
		\centering
		\setlength{\tabcolsep}{1.80mm}{
				\begin{tabular}{c||cc||cc||cc||cc||cc||cc}
					\hline\hline
					\multirow{2}{*}{\begin{tabular}[c]{@{}c@{}}MR of\\       (A, V, L)\end{tabular}} & \multicolumn{2}{c||}{(0.8, 0.5, 0.5)}          & \multicolumn{2}{c||}{(0.5, 0.8, 0.5)}          & \multicolumn{2}{c||}{(0.5, 0.5, 0.8)}         & \multicolumn{2}{c||}{(0.2, 0.5, 0.8)}          & \multicolumn{2}{c||}{(0.5, 0.2, 0.8)}          & \multicolumn{2}{c}{(0.8, 0.5, 0.2)}          \\ \cline{2-13} 
					& \multicolumn{1}{c|}{M.N.}   & R.C.         & \multicolumn{1}{c|}{M.N.}   & R.C.         & \multicolumn{1}{c|}{M.N}   & R.C.        & \multicolumn{1}{c|}{M.N.}   & R.C.         & \multicolumn{1}{c|}{M.N.}   & R.C.         & \multicolumn{1}{c|}{M.N.}   & R.C.         \\ \hline\hline
					A                                                                                & \multicolumn{1}{c|}{0.5067} & 0.5129          & \multicolumn{1}{c|}{0.5505} & 0.5476          & \multicolumn{1}{c|}{0.5054} & 0.4922         & \multicolumn{1}{c|}{0.5260} & 0.5122          & \multicolumn{1}{c|}{0.4691} & 0.5033          & \multicolumn{1}{c|}{0.4842} & 0.4878          \\ \hline
					V                                                                                & \multicolumn{1}{c|}{0.5746} & 0.5199          & \multicolumn{1}{c|}{0.5965} & 0.4520          & \multicolumn{1}{c|}{0.5542} & 0.5206         & \multicolumn{1}{c|}{0.5771} & 0.5067          & \multicolumn{1}{c|}{0.6077} & 0.4995          & \multicolumn{1}{c|}{0.5554} & 0.4804          \\ \hline
					L                                                                                & \multicolumn{1}{c|}{0.5792} & 0.6086          & \multicolumn{1}{c|}{0.4639} & 0.5841          & \multicolumn{1}{c|}{0.2239} & 0.5879         & \multicolumn{1}{c|}{0.1223} & 0.5886          & \multicolumn{1}{c|}{0.1499} & 0.5905          & \multicolumn{1}{c|}{0.2619} & 0.5769          \\ \hline
					AV                                                                               & \multicolumn{1}{c|}{0.6534} & 0.6390          & \multicolumn{1}{c|}{0.6026} & 0.6062          & \multicolumn{1}{c|}{0.5603} & 0.6479         & \multicolumn{1}{c|}{0.5787} & 0.5752          & \multicolumn{1}{c|}{0.634}  & 0.6209          & \multicolumn{1}{c|}{0.6073} & 0.6319          \\ \hline
					AL                                                                               & \multicolumn{1}{c|}{0.6142} & 0.6654          & \multicolumn{1}{c|}{0.5643} & 0.6454          & \multicolumn{1}{c|}{0.3973} & 0.6185         & \multicolumn{1}{c|}{0.5201} & 0.5897          & \multicolumn{1}{c|}{0.2943} & 0.6462          & \multicolumn{1}{c|}{0.3102} & 0.6413          \\ \hline
					VL                                                                               & \multicolumn{1}{c|}{0.6453} & 0.6867          & \multicolumn{1}{c|}{0.6567} & 0.6334          & \multicolumn{1}{c|}{0.2283} & 0.6669         & \multicolumn{1}{c|}{0.5577} & 0.6573          & \multicolumn{1}{c|}{0.1673} & 0.6500          & \multicolumn{1}{c|}{0.3020} & 0.6371          \\ \hline
					AVL                                                                              & \multicolumn{1}{c|}{0.6328} & 0.7285          & \multicolumn{1}{c|}{0.6343} & 0.6768          & \multicolumn{1}{c|}{0.4230} & 0.7051         & \multicolumn{1}{c|}{0.5718} & 0.6781          & \multicolumn{1}{c|}{0.3629} & 0.6977          & \multicolumn{1}{c|}{0.3257} & 0.6961          \\ \hline
					Ave.                                                                             & \multicolumn{1}{c|}{0.6088} & \textbf{0.6306} & \multicolumn{1}{c|}{0.5986} & \textbf{0.6088} & \multicolumn{1}{c|}{0.4748} & \textbf{0.618} & \multicolumn{1}{c|}{0.5347} & \textbf{0.5999} & \multicolumn{1}{c|}{0.4514} & \textbf{0.6109} & \multicolumn{1}{c|}{0.4393} & \textbf{0.6017} \\ \hline\hline
				\end{tabular}

	 }
	\vspace{-0.2cm}
\caption{F1 score of RedCore (R.C.) and MMIN (M.N.) with imbalanced missing rates (MR) for modalities (A, V, L).} \label{tab:MNRC}
\vspace{-0.6cm}
\end{table*}

\begin{figure*}[ht]
	\centering
	\vspace{-0.1cm}
	\begin{minipage}[b]{0.333\linewidth} 
		\centering
		\includegraphics[height=1.72in]{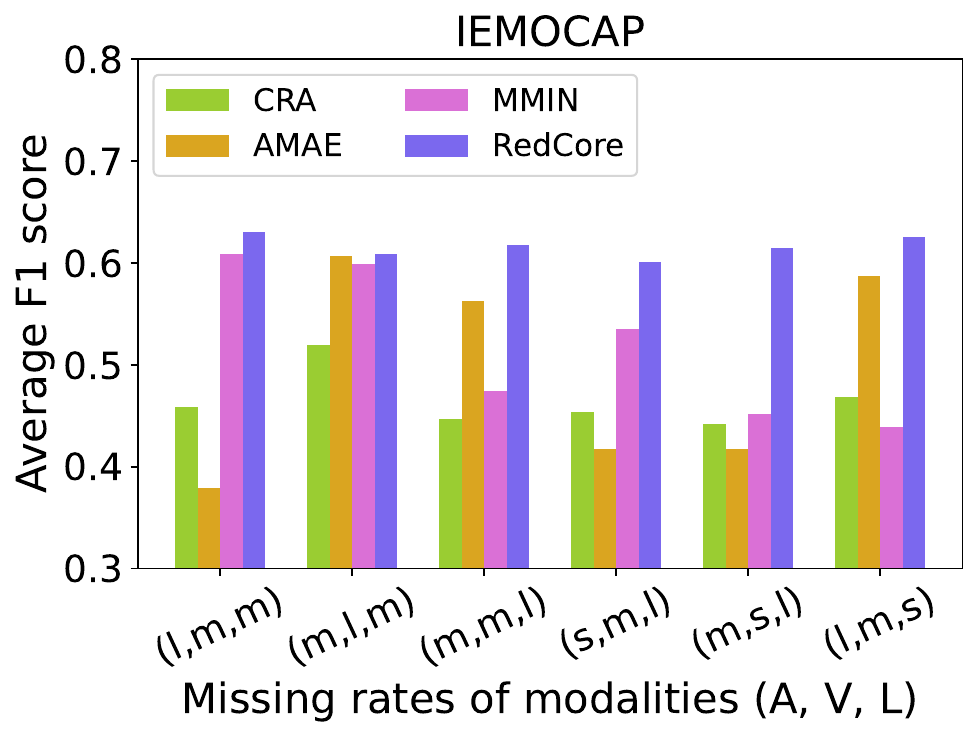}  
		\leftline{~~~(a) {\small F1} score vs missing rate on {\small IEMOCAP.}}\medskip
		\label{fig:side:a}
	\end{minipage}%
	\begin{minipage}[b]{0.333\linewidth} 
		\centering
		\includegraphics[height=1.72in]{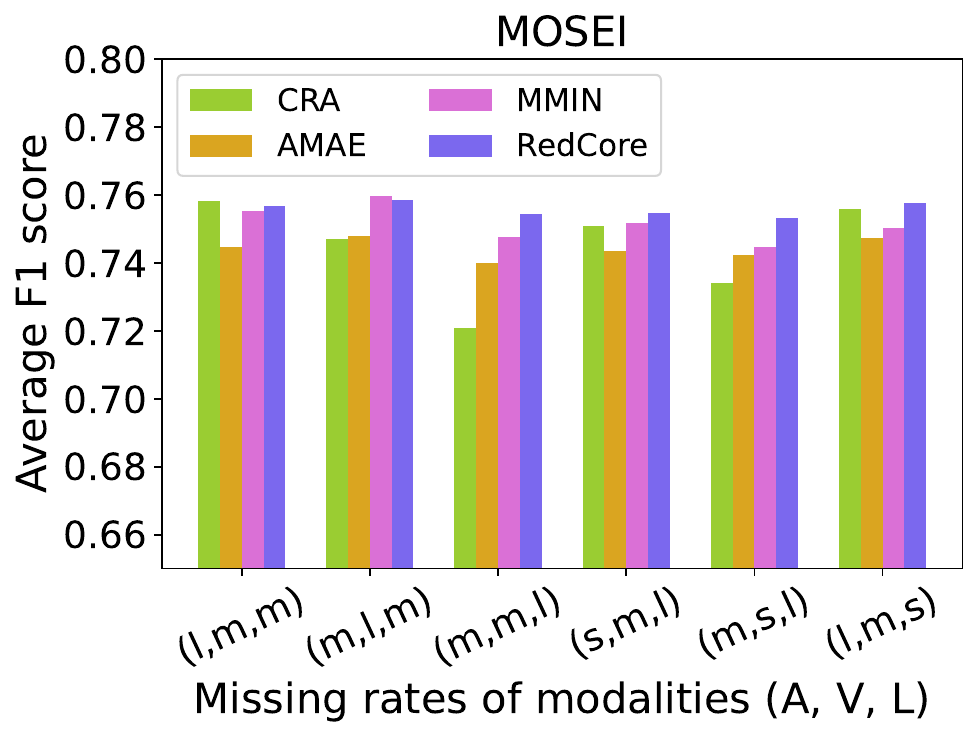}
		\centerline{~~~~~~~~(b) {\small F1} score vs missing rate on {\small MOSEI.}}\medskip
		\label{fig:side:b}
	\end{minipage}%
	\begin{minipage}[b]{0.333\linewidth} 
		\centering
		\includegraphics[height=1.72in]{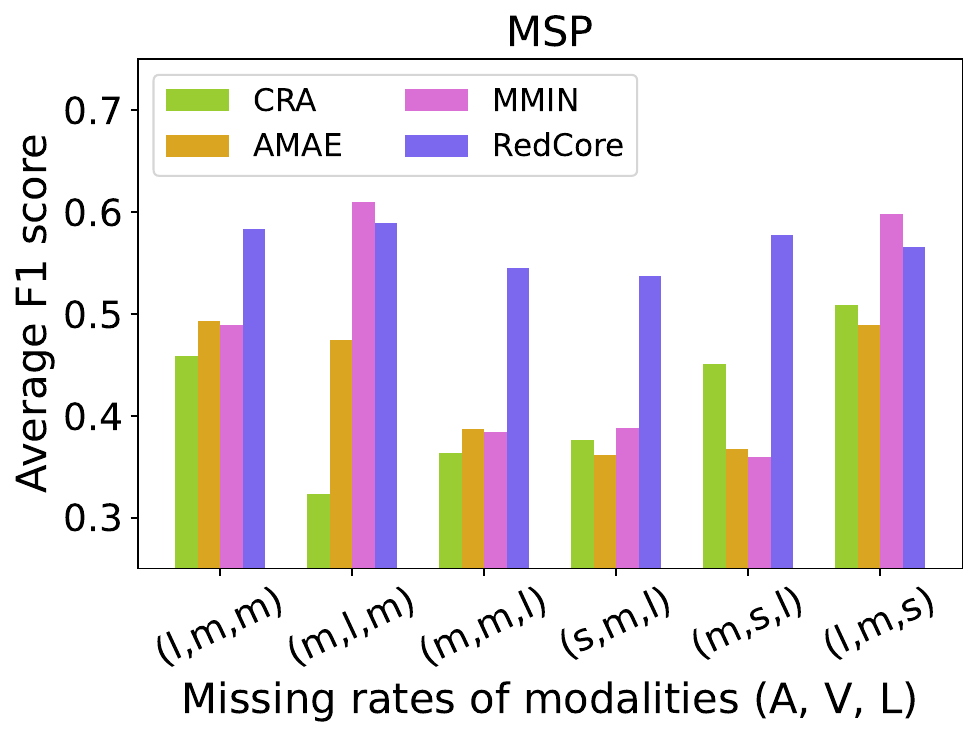}
		\centerline{~~~(c) {\small F1} score vs missing rate on {\small MSP.}}\medskip
		\label{fig:side:c}
	\end{minipage}%
	\vspace{-0.5cm}
	\caption{Average F1 score with imbalanced missing rates for different datasets (l=0.8, m=0.5, and s=0.2).}\label{fig:F1vsMRvar}
	\vspace{-0.2cm}
\end{figure*}

\vspace{0.15cm}
\noindent\textbf{{C. Relative advantage aware supervision regulation}}
\vspace{0.1cm}

\noindent The optimizer for training is usually based on stochastic gradient, and consequently, the loss might fluctuate significantly with iterations. To alleviate the fluctuation, we use the following moving average loss (MAL) to replace $\mathcal{L}_{M\!S\!E}^m(\bm{\theta})$ in eqs.~\eqref{eq:aveloss} and \eqref{eq:ra} for relative advantage calculation.
\begin{equation} \label{eq:mal}
	\mathcal{L}_{M\!A\!L}^m(\bm{\theta}_k)= (1-\beta)\mathcal{L}_{M\!A\!L}^m(\bm{\theta}_{k-1}) + \beta \mathcal{L}_{M\!S\!E}^m(\bm{\theta}_{k}),
\end{equation}
where $k$ represents the index of $\bm{\theta}$ updating step, and $\beta\in(0,1]$ is a pre-specified constant.
Following the work \cite{yang2021provably}, we propose a double-loop algorithm to solve the bi-level optimization problem, as summarized in Algorithm~\ref{alg:red}. The supervision $\bm{\eta}$ is iteratively updated for every $K$-step updates of the model parameter $\bm{\theta}$.

\section{Numerical Results}
\noindent \textbf{Benchmark datasets and Models:} In this section we evaluate RedCore on three commonly used multimodal datasets: IEMOCAP \cite{busso2008iemocap}, CMU-MOSEI \cite{zadeh2018multimodal} and MSP-IMPROV \cite{busso2016msp}. All these datasets include three modalities: acoustic (A), visual (V) and lexical (L). IEMOCAP and  MSP-IMPROV in our experiments are for four-class emotion recognition, and CMU-MOSEI is for sentiment analysis. For datasets IEMOCAP and MSP-IMPROV, we employ the same feature extraction method as \cite{zhao2021missing} to obtain sequential feature vectors for each modality. We utilize multimodal sentiment analysis feature extraction toolkit \cite{mao2022m} to attain the feature vectors for CMU-MOSEI.
In the comparison studies, the baseline models are cascaded residual autoencoder (CRA) \cite{zhao2021missing}, action masked autoencoder (AMAE) \cite{woo2023towards} and missing modality imagination network (MMIN) \cite{zhao2021missing}. More details about the benchmark datasets and models are provided in the supplementary materials.

\begin{figure*}[ht]
	\centering
	\vspace{-0.1cm}
	\begin{minipage}[b]{0.34\linewidth} 
		\centering
		\includegraphics[height=1.68in]{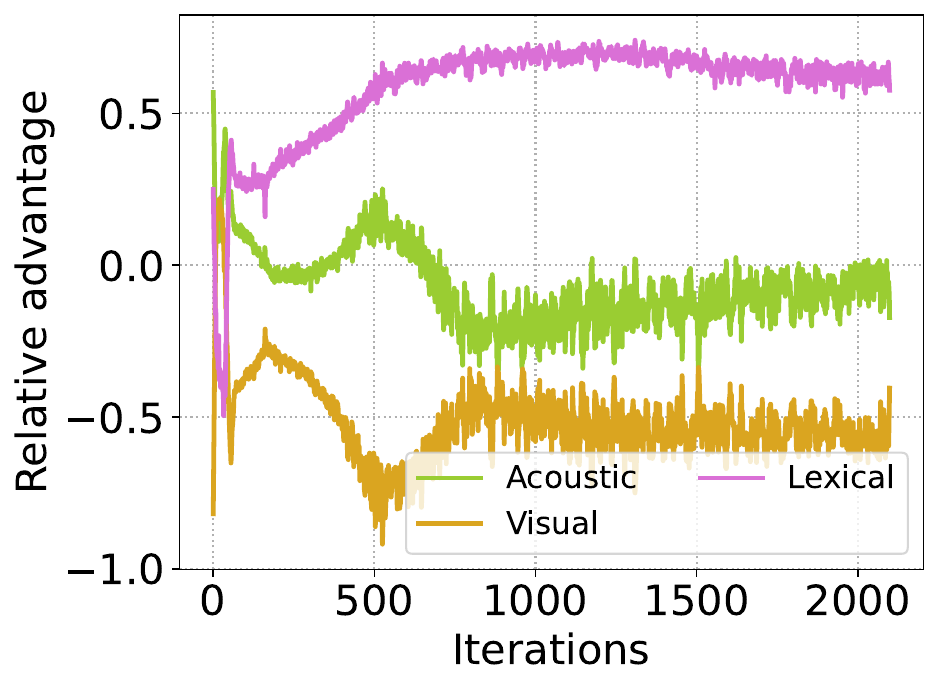}  
		\centerline{(a) Raletive advantage of Core.}\medskip
		\label{fig:side:a}
	\end{minipage}%
	\begin{minipage}[b]{0.34\linewidth} 
		\centering
		\includegraphics[height=1.68in]{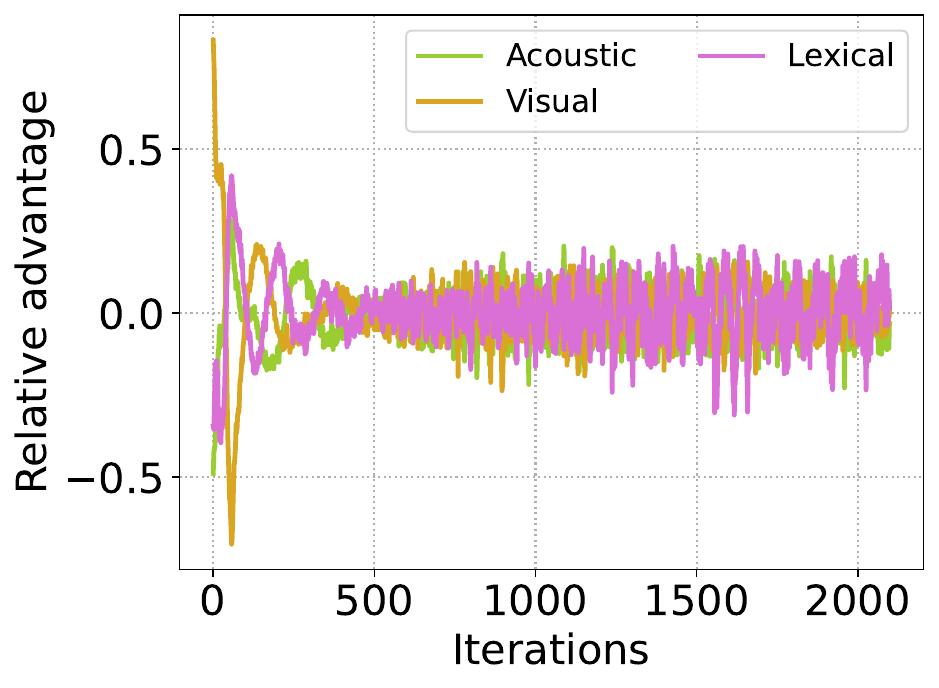}
		\centerline{~~~~(b) Raletive advantage of RedCore.}\medskip
		\label{fig:side:b}
	\end{minipage}%
	\begin{minipage}[b]{0.32\linewidth} 
		\centering
		\includegraphics[height=1.68in]{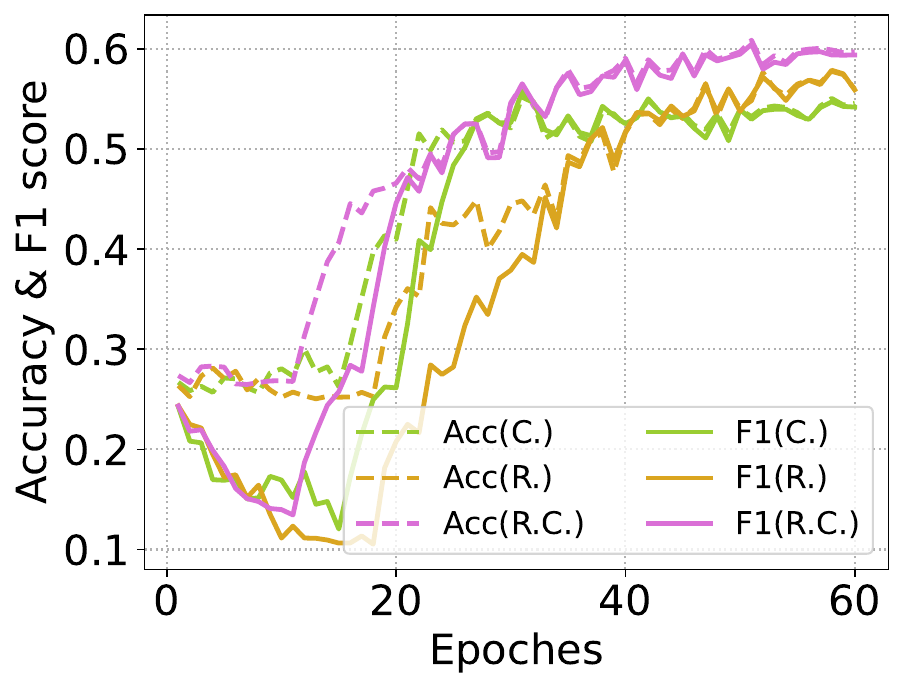}
		\centerline{(c) Learning curves.}\medskip
		\label{fig:side:c}
	\end{minipage}%
	\vspace{-0.4cm}
	\caption{Comparisons of Core, Red and RedCore on dataset IEMOCAP.}\label{fig:abla}
	\vspace{-0.5cm}
\end{figure*}

\subsection{Implementation Details}

 We employ transformer encoders as self-model encoders (senc) for all modalities, residual autoencoders as cross-modal encoders (xenc) and multilayer perceptrons (MLP) as decoders (dec). The number of self-attention layers for all self-modal encoders (senc) is 3, and all layers are with 8 heads and 128-dimensional hidden embedding. The cross-modal encoders for all modalities consist of 5 residual autoencoder blocks with RA-layers of size 256-128-64-32-64-128-256, and the outputs are 128-dimensional vectors as the representation for corresponding modalities. The coefficient $\gamma$ for the VIB loss is set to be 0.008. Optimizer Adam with learning rate $2\times10^{-4}$, momentum coefficient $(0.9, 0.999)$ and batch size 128 is adopted for model training. The parameter settings for supervision regulation are $p=2, \xi_1 = 0.015, \xi_2 = 0.15, K=3, \alpha_1 = 0.1, \beta=0.7$.  More details of the implementation can be found from the code in the supplementary materials. To mimic modality missing, the modalities are randomly masked according to their missing rates. The competing models adopt perfect data training (PDT), meaning the ``missing data" is used indirectly for training. In contrast, RedCore applies imperfect data training (IDT), i.e., the missing data is inaccessible. The experiments are carried out on 4 Nvidia Tesla V100 GPUs with 32G memory. Throughout this section, we use weighted F1 score as the performance metric. The reported results are obtained by averaging over 3 repeated experiments.

\subsection{Comparison Studies}
In the comparative analysis, we evaluate the models with two settings --- the training data with the same missing rate for all modalities, and with imbalanced missing rates. 

\noindent \textbf{Missing Modalities with the same missing rate:}
The models are trained on three datasets with shared missing rates ranging from 0.0 to 0.9, and are tested with all combinations of input modalities (i.e., A, V, L, AV, AL, VL, AVL). Detailed results of the F1 score for each input combination are reported by the table in the supplementary materials. Figure~\ref{fig:F1vsMR} illustrates the average F1 score (average over all the input combinations) on the three datasets. 
  Three observations about all tested datasets from Figure~\ref{fig:F1vsMR} demonstrate the superiority of RedCore over the competing models. 1) When the missing rate is low, i.e., 0.0 and 0.1, RedCore outperforms all the baseline models. This is because other compared models are not fully exposed to modality missing, and hence have difficulty in coping with the situation, while RedCore consistently conducts cross-modal representation learning even when no modality is missing. 2) when the missing rate is at a moderate level, (e.g., between 0.2 and 0.6 on IEMOCAP), RedCore shows at least comparable performance with other models, despite adopting IDT training regime with less data. This verifies that RedCore is more data-efficient than the baseline models. 3) when the missing rate is large, (e.g., between 0.7 and 0.9 on IEMOCAP), RedCore can still maintain the performance or have slight to moderate performance drop, whereas the compared models all show significant performance degradation. This observation validates the exploitation of supervision from the available modalities and the labels boosts the performance of RedCore.
   In conclusion, RedCore has demonstrated superior robustness over a wide spectrum of missing rates.

\begin{table}[t]
	\centering
	\small
	\setlength{\tabcolsep}{1.7mm}{
		\begin{tabular}{l||l|l|l|l|l|l}
			\hline\hline
			MR   & (s,m,l) & (m,s,l) & (l,m,s) & (l,s,m) & (s,l,m) & (m,l,s) \\ \hline\hline
			C.   & 0.5696  & 0.5599  & 0.5653  & 0.5534  & 0.5481  & 0.5651  \\ \hline
			R.   & 0.5793  & 0.5708  & 0.5894  & 0.5831  & 0.5853  & 0.5849  \\ \hline
			R.C. & 0.5999  & 0.6109  & 0.6017  & 0.5921  & 0.5993  & 0.5950  \\ \hline\hline
	\end{tabular} }
	\vspace{-0.3cm}
	\caption{Average F1 score with imbalanced missing rates (MR) for modalities (A,V,L) (l=0.8, m=0.5, and s=0.2).}
	\label{tab:abla}
	\vspace{-0.1cm}
\end{table}

\noindent \textbf{Missing modalities with imbalanced missing rates:} Table~\ref{tab:MNRC} compares MMIN and RedCore trained with imbalanced missing rates and reports the average F1 score of different input combinations. Comparing the results on missing rates (0.8, 0.5, 0.5) and (0.8, 0.5, 0.2) (or (0.5, 0.5, 0.8) and (0.2, 0.5, 0.8)), it is clear that both MMIN and RedCore exhibit better performance with the former, even though the former provides less data. This confirms that more imbalanced missing rates (the latter) can lead to deteriorated performance, and MMIN is more sensitive to the imbalance. 
 It is also shown that MMIN's lexical modality is sensitive to imbalanced modality missing. In comparison, RedCore showcases greater robustness against imbalanced missing rates. This is further supported by Figure~\ref{fig:F1vsMRvar} showing the average F1 score of all the baseline models on the three datasets.

\subsection{Ablation Studies}
In the ablation studies, we compare RedCore with two ablative settings on dataset IEMOCAP: Core --- the Red part is excluded, namely, $\eta^1=\eta^2=\eta^3$; and Red --- the Core part is partially removed, namely, VIB based individual modality label prediction is banned during training, and only the joint modality conducts prediction (the second term in eq.~\eqref{eq:LVIB} only accounts for $m=M+1$ instead of summing over $m$ from $1$ to $M+1$). The models are trained with imbalanced missing rates and tested with all possible input modality combinations, and the average F1 score is reported in  Table~\ref{tab:abla}. Via comparing Core, Red and RedCore, it is obvious that VIB based cross-modal representation learning and relative advantage aware supervision regulation both contribute to the success of RedCore. (Due to the limited space in the tables and figures, we use C., R., and R.C. as the shorthand for Core, Red, and RedCore, respectively).

Figures~\ref{fig:abla}(a) and \ref{fig:abla}(b) plot the relative advantages of the three modalities during training with missing rates (0.8, 0.2, 0.5) for Core and RedCore, respectively. Figure~\ref{fig:abla}(a) shows that lexical modality dominates the other two when no supervision regulation is enforced. Comparing Figures~\ref{fig:abla}(a) and \ref{fig:abla}(b), it is clear that supervision regulation achieves more balanced learning, indicating a smaller imputation loss gap between modalities. Shown in Figure~\ref{fig:abla}(c) are the learning curves (accuracy and F1 score) of the three settings, which demonstrates that RedCore is improved over Core and Red.

\section{Conclusions}
In this paper, we propose a cross-modal representation learning approach for missing modalities based on the VIB theory. Considering the imbalanced missing rates, a bi-level optimization problem is formulated to dynamically regulate the supervision of each modality during training. The numerical results validate the effectiveness of our design, and demonstrate the  robustness of the proposed approach against modality missing. In our future work, we will investigate the convergence of algorithm~\ref{alg:red} and analyze the Pareto optimality of the solution according to Lemma~\ref{lemma:convex}.

\bibliography{aaai23}

\newpage
\section*{Supplementary Materials for Paper ``RedCore: Relative Advantage Aware Cross-modal Representation Learning for Missing Modalities with Imbalanced Missing Rates"}
The numbers of the figures and table in the supplementary materials follow those in the paper. 
\subsection{A. Proof of Lemma~\ref{lemma:convex}.}
\begin{proof}
	It follows from the definition of $\bm{r}(\bm{\theta})$ that:
	\begin{equation}
		\bm{r}(\bm{\theta})^T\bm{1} = 0.
	\end{equation}
	Thus, for non-trial case, i.e., $\bm{r}(\bm{\theta})\neq\bm{0}$, there exists an index $i\in[M]$, such that $RA^i(\bm{\theta})<0$.

	We define $\Gamma_1=\{\bm{\eta}\in\mathbb{R}^M\mid\bm{\eta} \ge \xi_1 \bm{1} , ||\bm{\eta}||_p = \xi_2\}$ and $\Gamma_2=\{\bm{\eta}\in\mathbb{R}^M\mid\bm{\eta} \ge \xi_1 \bm{1}, ||\bm{\eta}||_p \le \xi_2\}$ to denote the feasible domains of problems \eqref{eq:bilevel} and \eqref{eq:bilevel2} with $\bm{\theta}$ fixed, respectively. It is evident that $\Gamma_1$ is non-convex due to the nonlinear equality constraint \eqref{eq:ub}, and $\Gamma_2$ is convex. Lemma~\ref{lemma:convex} can be interpreted as the optimal solution to problem \eqref{eq:bilevel2} lies in $\Gamma_1$, which will be proved by contradiction in the following.
	
	Assume $\bm{\eta}^* \in \Gamma_2$ is an optimal solution to problem \eqref{eq:bilevel2} and $\bm{\eta}^* \notin \Gamma_1$, which means $\bm{\eta}^*$ is an inner point of $\Gamma_2$. Consequently, there exists a sufficiently small positive constant $\delta>0$ such that $\bm{\eta}' = \bm{\eta}^* + \delta\cdot\bm{1}_i \in \Gamma_2$, where $\bm{1}_i$ denotes a vector in which the $i$-th element is 1 and all other elements are 0. Given that $RA^i(\bm{\theta})<0$, we have $\bm{r}(\bm{\theta})^T\bm{\eta}' < \bm{r}(\bm{\theta})^T\bm{\eta}^*$, indicating that $\bm{\theta}^*$ is not the optimal solution. This contradicts the initial assumption and concludes the proof.
\end{proof}

\subsection{B. Derivation of $\tilde{\mathcal{L}}_{V\!I\!B}(\bm{\theta})$}

The variational encoders take the Gaussian form $p(\bm{z}^m|\bm{x}^m; \bm{\theta}^m_{enc}) = \mathcal{N}(\bm{z}^m; f_{\bm{\mu}}(\bm{x}^m;\bm{\theta}^m_{enc}), f_{\bm{\Sigma}}(\bm{x}^m;\bm{\theta}^m_{enc}))$, where $f_{\bm{\mu}}(\cdot; \bm{\theta}^m_{enc})\in\mathbb{R}^d$ and  $f_{\bm{\Sigma}}(\cdot; \bm{\theta}^m_{enc})\in\mathbb{R}^{d\times d}$ are the mean vector and covariance matrix determined by the network models as will be specified in the Numerical Results section. $q(\bm{z}^m)$ is fixed as a $d$-dimensional spherical Gaussian, that is $q(\bm{z}^m)=\mathcal{N}(\bm{z}^m; \bm{0}_d, \bm{I}_d)$, where $\bm{0}_d$ and $\bm{I}_d$ are $d$-dimensional all-zero vector and $d\times d$ identity matrix, respectively. The decoders are simple classifier models of the form $q(y^m|\bm{z}^m) =\text{softmax}(f_c(\bm{z}^m))$, where $f_c(\bm{z}^m), m\in[M+1]$ are MLPs that map the representations to the logits for classification. Applying the reparameterization trick \cite{kingmaauto}, we obtain $\bm{z}^m$ as follows:
\begin{equation*}
	\bm{z}^m = f_{\bm{\mu}}(\bm{x}^m;\bm{\theta}^m_{enc}) + \bm{\epsilon} \odot f_{\bm{\Sigma}}(\bm{x}^m;\bm{\theta}^m_{enc})), 
\end{equation*}
where $\odot$ means element-wise product and $\bm{\epsilon}\sim \mathcal{N}(\bm{0}_d, \bm{I}_d)$.
As a result, $\mathcal{L}_{V\!I\!B}(\bm{\theta})$ can be approximated by $\tilde{\mathcal{L}}_{V\!I\!B}(\bm{\theta})$:
\begin{equation*}
	\begin{aligned}
		\!\!\! \tilde{\mathcal{L}}_{V\!I\!B}(\bm{\theta}) =  & \frac{1}{N} \sum\limits_{n=1}^N\sum\limits_{m=1}^{M+1} D_{K\!L}(p(\bm{z}_n^m|\bm{x}_n^m;\bm{\theta}^m_{enc})||q(\bm{z}_n^m)) \\
		&  ~~~~~~~~~~~~~~~~~~~~ -\gamma \mathbb{E}_{\bm{\epsilon}}[\log q(y_n^m|\bm{z}_n^m; \bm{\theta}^m_{dec})].
	\end{aligned}
\end{equation*}
This formulation circumvents the expectation in eq.~\eqref{eq:LVIB}, and thus allows us to compute unbiased stochastic gradient through backpropagating of a single sample.

\subsection{C. Benchmark Datasets and Baseline Models}

\noindent \textbf{Benchmark datasets:} IEMOCAP contains conversation videos where two actors perform improvised or scripted scenarios, specially selected to evoke emotional expressions. The conversations are divided into utterances which are annotated into categorical emotion labels. Following the popular label processing method as in work \cite{liang2020semi}, we obtain the four-class dataset.
MSP-IMPROV consist of data from six dyadic conversations involving 12 actors. For a fair comparison, we adopt the same video selection scheme described in work \cite{zhao2021missing} to set up the four-class emotion recognition dataset.
MOSEI is a collection of over 23,500 sentence utterance videos from more than 1000 online YouTube speakers. 
Each utterance is annotated with sentiment labels: positive, negative and neutral. The sentiment score is given between $[-3, 3]$. For datasets IEMOCAP and MSP-IMPROV, we employ the same feature extraction method as \cite{zhao2021missing} to obtain sequential feature vectors for each modality. We utilize multimodal sentiment analysis feature extraction toolkit \cite{mao2022m} to attain the sequential feature vectors for CMU-MOSEI.

\noindent \textbf{Baseline models:} As an extension of autoencoder (AE), CRA stacks a series of residual AE to form a cascaded architecture for missing modality imputation. AMAE is originally designed for robust action recognition with visual modalities (RGB, depth and infrared), and we adapt it to our scenario with acoustic, visual and lexical modalities.  MMIN is a strong baseline model that combines CRA and cycle consistency learning for the representation learning of missing modalities.  

\subsection{D. Model Architecture}
The proposed model architecture is visualized in Figure~\ref{fig:traintest}, where Figure~\ref{fig:traintest}(a) shows the training computation flow on a sample with 3 complete modalities, Figures~\ref{fig:traintest}(b) and \ref{fig:traintest}(c) the training and inference with modality 2 missing.

\subsection{E. Performance Comparison with Baseline Models on the Three Datasets}
Table~\ref{tab:comp4models} reports the results of all compared models on three datasets with shared missing rates ranging form 0.0 to 0.9 for training. For dataset IEMOCAP, we show the test results of different input modality combinations, and the Ave. column is the average F1 score of all combinations. We only exhibit the average F1 score for the other two datasets. From the table, it can be concluded that MMIN and Redcore are roughly the two leading models. 

\subsection{F. Code and Data}
Our code has been submitted as supplementary material for review, and will be released with the publication of the paper. The code and the features extracted from the three datasets are also available at: \url{https://github.com/nobodyresearcher/RedCore}.

\begin{figure*}[t]
	\centering
	\begin{minipage}[b]{0.33\linewidth} 
		\centering
		\includegraphics[width=2.15in]{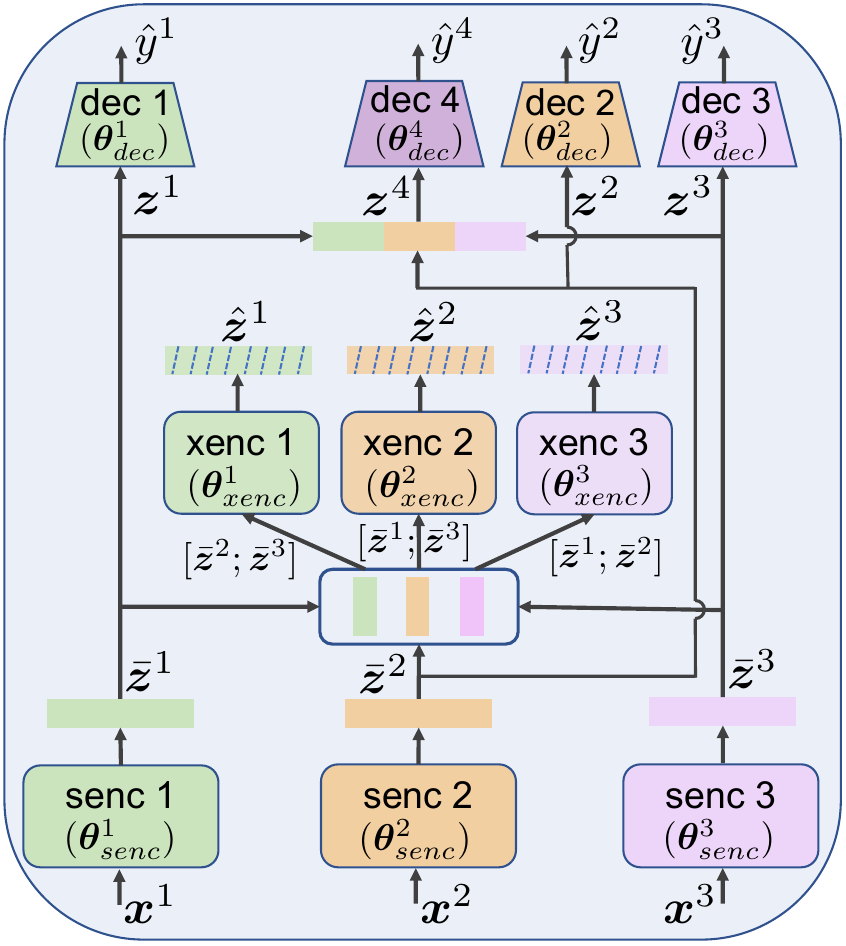}
		\centerline{(a)}\medskip
		\label{fig:side:a}
	\end{minipage}%
	\begin{minipage}[b]{0.33\linewidth} 
		\centering
		\includegraphics[width=2.15in]{frametrmiss0807.pdf}
		\centerline{(b)}\medskip
		\label{fig:side:b}
	\end{minipage}%
	\begin{minipage}[b]{0.33\linewidth} 
		\centering
		\includegraphics[width=2.15in]{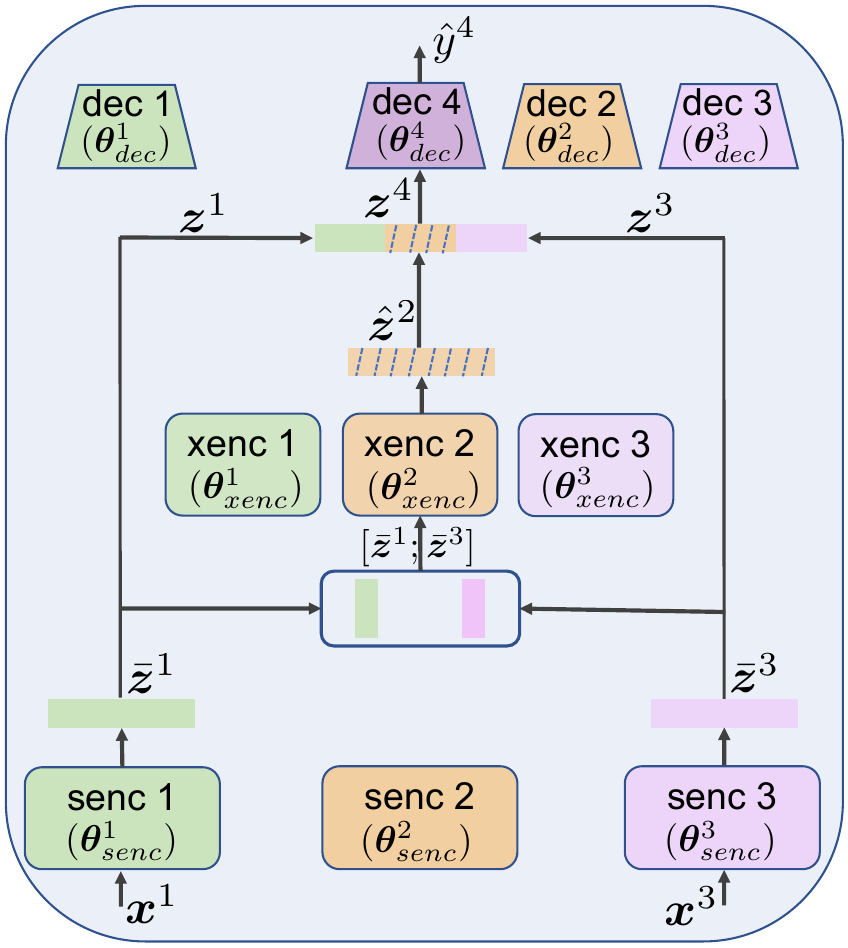}
		\centerline{(c)}\medskip
		\label{fig:side:c}
	\end{minipage}%
	\vspace{-0.4cm}
	\caption{Model architecture (with 3 modalities as an example). (a) Training on a sample with all modalities available. (b) Training on a sample with modality 2 missing. (c) Inference on a sample with modality 2 missing.}\label{fig:traintest_all}
	\vspace{-0.5cm}
\end{figure*}

\begin{table*}
		\small
	\centering
	\begin{tabular}{cc||cccccccc||c||c}
		\hline\hline
		\multicolumn{2}{c||}{\multirow{2}{*} {\diagbox{MRT}{TMC} }}                        & \multicolumn{8}{c||}{IEMOCAP}                                                                                                                                                                                                                                                                     & MOSEI           & MSP             \\ \cline{3-12} 
		\multicolumn{2}{c||}{}                                        & \multicolumn{1}{c|}{A}               & \multicolumn{1}{c|}{V}               & \multicolumn{1}{c|}{L}               & \multicolumn{1}{c|}{AV}              & \multicolumn{1}{c|}{AL}              & \multicolumn{1}{c|}{VL}              & \multicolumn{1}{c|}{AVL}             & Ave.            & Ave.            & Ave.            \\ \hline\hline
		\multicolumn{1}{c|}{\multirow{4}{*}{0.0}}   & CRA              & \multicolumn{1}{c|}{0.065}           & \multicolumn{1}{c|}{0.2422}          & \multicolumn{1}{c|}{0.134}           & \multicolumn{1}{c|}{0.1297}          & \multicolumn{1}{c|}{0.1363}          & \multicolumn{1}{c|}{0.2223}          & \multicolumn{1}{c|}{0.1403}          & 0.2165          & 0.6739          & 0.3162          \\ \cline{2-12} 
		\multicolumn{1}{c|}{}                     & AMAE             & \multicolumn{1}{c|}{0.4235}          & \multicolumn{1}{c|}{0.4146}          & \multicolumn{1}{c|}{0.385}           & \multicolumn{1}{c|}{0.5449}          & \multicolumn{1}{c|}{0.6132}          & \multicolumn{1}{c|}{0.4859}          & \multicolumn{1}{c|}{0.6814}          & 0.5475          & 0.74479         & 0.5312          \\ \cline{2-12} 
		\multicolumn{1}{c|}{}                     & MMIN             & \multicolumn{1}{c|}{0.3255}          & \multicolumn{1}{c|}{0.3451}          & \multicolumn{1}{c|}{0.3656}          & \multicolumn{1}{c|}{0.5334}          & \multicolumn{1}{c|}{0.592}           & \multicolumn{1}{c|}{0.4621}          & \multicolumn{1}{c|}{0.692}           & 0.531           & 0.7414          & 0.5325          \\ \cline{2-12} 
		\multicolumn{1}{c|}{}                     & \textbf{RedCore} & \multicolumn{1}{c|}{\textbf{0.5411}} & \multicolumn{1}{c|}{\textbf{0.3638}} & \multicolumn{1}{c|}{\textbf{0.6034}} & \multicolumn{1}{c|}{\textbf{0.6387}} & \multicolumn{1}{c|}{\textbf{0.6929}} & \multicolumn{1}{c|}{\textbf{0.6391}} & \multicolumn{1}{c|}{\textbf{0.7489}} & \textbf{0.6284} & \textbf{0.7551} & \textbf{0.6004} \\ \hline\hline
		\multicolumn{1}{c|}{\multirow{4}{*}{0.1}} & CRA              & \multicolumn{1}{c|}{0.4548}          & \multicolumn{1}{c|}{0.3984}          & \multicolumn{1}{c|}{0.5461}          & \multicolumn{1}{c|}{0.5725}          & \multicolumn{1}{c|}{0.6129}          & \multicolumn{1}{c|}{0.6147}          & \multicolumn{1}{c|}{0.6501}          & 0.57            & 0.7584          & 0.5622          \\ \cline{2-12} 
		\multicolumn{1}{c|}{}                     & AMAE             & \multicolumn{1}{c|}{0.4815}          & \multicolumn{1}{c|}{0.4617}          & \multicolumn{1}{c|}{0.5618}          & \multicolumn{1}{c|}{0.609}           & \multicolumn{1}{c|}{0.6286}          & \multicolumn{1}{c|}{0.6423}          & \multicolumn{1}{c|}{0.6816}          & 0.597           & 0.74429         & 0.5963          \\ \cline{2-12} 
		\multicolumn{1}{c|}{}                     & MMIN             & \multicolumn{1}{c|}{0.4628}          & \multicolumn{1}{c|}{0.4288}          & \multicolumn{1}{c|}{0.6092}          & \multicolumn{1}{c|}{0.6305}          & \multicolumn{1}{c|}{0.6635}          & \multicolumn{1}{c|}{0.6674}          & \multicolumn{1}{c|}{0.7151}          & 0.619           & 0.7484          & 0.6006          \\ \cline{2-12} 
		\multicolumn{1}{c|}{}                     & \textbf{RedCore} & \multicolumn{1}{c|}{\textbf{0.5267}} & \multicolumn{1}{c|}{\textbf{0.4309}} & \multicolumn{1}{c|}{\textbf{0.5955}} & \multicolumn{1}{c|}{\textbf{0.6373}} & \multicolumn{1}{c|}{\textbf{0.671}}  & \multicolumn{1}{c|}{\textbf{0.6542}} & \multicolumn{1}{c|}{\textbf{0.7212}} & \textbf{0.6203} & \textbf{0.758}  & \textbf{0.598}  \\ \hline\hline
		\multicolumn{1}{c|}{\multirow{4}{*}{0.3}} & CRA              & \multicolumn{1}{c|}{0.4797}          & \multicolumn{1}{c|}{0.5287}          & \multicolumn{1}{c|}{0.5215}          & \multicolumn{1}{c|}{0.5834}          & \multicolumn{1}{c|}{0.6107}          & \multicolumn{1}{c|}{0.5967}          & \multicolumn{1}{c|}{0.6439}          & 0.5788          & 0.7589          & 0.597           \\ \cline{2-12} 
		\multicolumn{1}{c|}{}                     & AMAE             & \multicolumn{1}{c|}{0.5376}          & \multicolumn{1}{c|}{0.5099}          & \multicolumn{1}{c|}{0.5899}          & \multicolumn{1}{c|}{0.5708}          & \multicolumn{1}{c|}{0.6314}          & \multicolumn{1}{c|}{0.6454}          & \multicolumn{1}{c|}{0.6725}          & 0.6001          & 0.74519         & 0.5965          \\ \cline{2-12} 
		\multicolumn{1}{c|}{}                     & MMIN             & \multicolumn{1}{c|}{0.5381}          & \multicolumn{1}{c|}{0.5489}          & \multicolumn{1}{c|}{0.5929}          & \multicolumn{1}{c|}{0.6253}          & \multicolumn{1}{c|}{0.6353}          & \multicolumn{1}{c|}{0.6915}          & \multicolumn{1}{c|}{0.6877}          & 0.6266          & 0.7478          & 0.5813          \\ \cline{2-12} 
		\multicolumn{1}{c|}{}                     & \textbf{RedCore} & \multicolumn{1}{c|}{\textbf{0.5217}} & \multicolumn{1}{c|}{\textbf{0.4623}} & \multicolumn{1}{c|}{\textbf{0.5763}} & \multicolumn{1}{c|}{\textbf{0.6123}} & \multicolumn{1}{c|}{\textbf{0.6545}} & \multicolumn{1}{c|}{\textbf{0.6488}} & \multicolumn{1}{c|}{\textbf{0.7215}} & \textbf{0.6167} & \textbf{0.7564} & \textbf{0.6069} \\ \hline\hline
		\multicolumn{1}{c|}{\multirow{4}{*}{0.5}} & CRA              & \multicolumn{1}{c|}{0.4349}          & \multicolumn{1}{c|}{0.5583}          & \multicolumn{1}{c|}{0.5766}          & \multicolumn{1}{c|}{0.5969}          & \multicolumn{1}{c|}{0.583}           & \multicolumn{1}{c|}{0.6464}          & \multicolumn{1}{c|}{0.6502}          & 0.5889          & 0.7519          & 0.5856          \\ \cline{2-12} 
		\multicolumn{1}{c|}{}                     & AMAE             & \multicolumn{1}{c|}{0.5021}          & \multicolumn{1}{c|}{0.547}           & \multicolumn{1}{c|}{0.5596}          & \multicolumn{1}{c|}{0.631}           & \multicolumn{1}{c|}{0.6216}          & \multicolumn{1}{c|}{0.6409}          & \multicolumn{1}{c|}{0.687}           & 0.6053          & 0.74759         & 0.5509          \\ \cline{2-12} 
		\multicolumn{1}{c|}{}                     & MMIN             & \multicolumn{1}{c|}{0.4975}          & \multicolumn{1}{c|}{0.5511}          & \multicolumn{1}{c|}{0.5856}          & \multicolumn{1}{c|}{0.5966}          & \multicolumn{1}{c|}{0.6014}          & \multicolumn{1}{c|}{0.6695}          & \multicolumn{1}{c|}{0.6725}          & 0.6063          & 0.7582          & 0.588           \\ \cline{2-12} 
		\multicolumn{1}{c|}{}                     & \textbf{RedCore} & \multicolumn{1}{c|}{\textbf{0.4968}} & \multicolumn{1}{c|}{\textbf{0.4772}} & \multicolumn{1}{c|}{\textbf{0.6142}} & \multicolumn{1}{c|}{\textbf{0.5942}} & \multicolumn{1}{c|}{\textbf{0.6367}} & \multicolumn{1}{c|}{\textbf{0.6783}} & \multicolumn{1}{c|}{\textbf{0.7071}} & \textbf{0.6135} &\textbf{0.7592}  & \textbf{0.592}\\ \hline\hline
		\multicolumn{1}{c|}{\multirow{4}{*}{0.7}} & CRA              & \multicolumn{1}{c|}{0.4466}          & \multicolumn{1}{c|}{0.2593}          & \multicolumn{1}{c|}{0.1}             & \multicolumn{1}{c|}{0.5693}          & \multicolumn{1}{c|}{0.4472}          & \multicolumn{1}{c|}{0.2756}          & \multicolumn{1}{c|}{0.3569}          & 0.4277          & 0.7465          & 0.316           \\ \cline{2-12} 
		\multicolumn{1}{c|}{}                     & AMAE             & \multicolumn{1}{c|}{0.5011}          & \multicolumn{1}{c|}{0.5669}          & \multicolumn{1}{c|}{0.5617}          & \multicolumn{1}{c|}{0.6254}          & \multicolumn{1}{c|}{0.5905}          & \multicolumn{1}{c|}{0.6126}          & \multicolumn{1}{c|}{0.6532}          & 0.5944          & 0.74059         & 0.4011          \\ \cline{2-12} 
		\multicolumn{1}{c|}{}                     & MMIN             & \multicolumn{1}{c|}{0.5025}          & \multicolumn{1}{c|}{0.553}           & \multicolumn{1}{c|}{0.3547}          & \multicolumn{1}{c|}{0.5695}          & \multicolumn{1}{c|}{0.3933}          & \multicolumn{1}{c|}{0.3581}          & \multicolumn{1}{c|}{0.4342}          & 0.5076          & 0.7574          & 0.4032          \\ \cline{2-12} 
		\multicolumn{1}{c|}{}                     & \textbf{RedCore} & \multicolumn{1}{c|}{\textbf{0.5277}} & \multicolumn{1}{c|}{\textbf{0.5182}} & \multicolumn{1}{c|}{\textbf{0.6251}} & \multicolumn{1}{c|}{\textbf{0.6405}} & \multicolumn{1}{c|}{\textbf{0.67}}   & \multicolumn{1}{c|}{\textbf{0.6865}} & \multicolumn{1}{c|}{\textbf{0.7227}} & \textbf{0.6393} & \textbf{0.7597} & \textbf{0.5796} \\ \hline\hline
		\multicolumn{1}{c|}{\multirow{4}{*}{0.9}} & CRA              & \multicolumn{1}{c|}{0.1294}          & \multicolumn{1}{c|}{0.1371}          & \multicolumn{1}{c|}{0.1369}          & \multicolumn{1}{c|}{0.1535}          & \multicolumn{1}{c|}{0.1089}          & \multicolumn{1}{c|}{0.2079}          & \multicolumn{1}{c|}{0.1003}          & 0.2459          & 0.6126          & 0.2447          \\ \cline{2-12} 
		\multicolumn{1}{c|}{}                     & AMAE             & \multicolumn{1}{c|}{0.4984}          & \multicolumn{1}{c|}{0.4537}          & \multicolumn{1}{c|}{0.1696}          & \multicolumn{1}{c|}{0.5865}          & \multicolumn{1}{c|}{0.1875}          & \multicolumn{1}{c|}{0.173}           & \multicolumn{1}{c|}{0.1836}          & 0.3887          & 0.5915          & 0.3095          \\ \cline{2-12} 
		\multicolumn{1}{c|}{}                     & MMIN             & \multicolumn{1}{c|}{0.5245}          & \multicolumn{1}{c|}{0.5125}          & \multicolumn{1}{c|}{0.1223}          & \multicolumn{1}{c|}{0.6039}          & \multicolumn{1}{c|}{0.2781}          & \multicolumn{1}{c|}{0.1223}          & \multicolumn{1}{c|}{0.2948}          & 0.4342          & 0.7136          & 0.3314          \\ \cline{2-12} 
		\multicolumn{1}{c|}{}                     & \textbf{RedCore} & \multicolumn{1}{c|}{\textbf{0.5313}} & \multicolumn{1}{c|}{\textbf{0.4126}} & \multicolumn{1}{c|}{\textbf{0.5681}} & \multicolumn{1}{c|}{\textbf{0.6658}} & \multicolumn{1}{c|}{\textbf{0.6195}} & \multicolumn{1}{c|}{\textbf{0.6374}} & \multicolumn{1}{c|}{\textbf{0.7077}} & \textbf{0.6145} & \textbf{0.7463} & \textbf{0.4824} \\ \hline\hline
	\end{tabular}
	\captionsetup{type=table}
	\vspace{-0.2cm}
	\caption{Performance (in terms of F1 score) comparison with baseline models (MRT and TMC stand for missing rate for training and tested modality combination, respectively).} 
	\label{tab:comp4models}
\end{table*}

\end{document}